\documentclass[letterpaper, 10 pt, conference]{ieeeconf}

\IEEEoverridecommandlockouts                          

\usepackage[table]{xcolor}
\usepackage{cite}
\usepackage{amsmath,amssymb,amsfonts}
\usepackage{algorithmic}
\usepackage{graphicx}
\usepackage{textcomp}
\usepackage{subcaption}
\usepackage{multirow}
\usepackage{booktabs}

\usepackage{academicons}
\usepackage{tikz}
\definecolor{orcidlogocol}{HTML}{A6CE39}
\definecolor{lime}{HTML}{A6CE39}
\DeclareRobustCommand{\orcidicon}{%
    \begin{tikzpicture}
    \draw[lime, fill=lime] (0,0) 
    circle [radius=0.16] 
    node[white] {{\fontfamily{qag}\selectfont \tiny ID}};
    \draw[white, fill=white] (-0.0625,0.095) 
    circle [radius=0.007];
    \end{tikzpicture}
    \hspace{-2mm}
}

\newcommand{\orcidMingyu}{\href{https://orcid.org/0000-0002-8752-7950}{\orcidicon}}
\newcommand{\orcidEkim}{\href{https://orcid.org/0000-0002-3103-6052}{\orcidicon}}
\newcommand{\orcidMarc}{\href{https://orcid.org/0009-0005-4987-9989}{\orcidicon}}
\newcommand{\orcidJun}{\href{https://orcid.org/0009-0006-2970-9816}{\orcidicon}}
\newcommand{\orcidWalter}{\href{https://orcid.org/0000-0003-4565-1272}{\orcidicon}}
\newcommand{\orcidXingcheng}{\href{https://orcid.org/0000-0003-1178-5221}{\orcidicon}}
\newcommand{\orcidLuka}{\href{https://orcid.org/0000-0001-5026-3368}{\orcidicon}}
\newcommand{\orcidYuning}{\href{https://orcid.org/0000-0002-1279-5539}{\orcidicon}}
\newcommand{\orcidKnoll
}{\href{https://orcid.org/0000-0003-4840-076X}{\orcidicon}}

\makeatletter
\let\NAT@parse\undefined
\makeatother

\usepackage{hyperref}
\hypersetup{
    bookmarks=false,
    colorlinks=true,
    breaklinks=true,
    linkcolor=blue,
    filecolor=magenta,      
    hypertexnames=true,
    urlcolor=cyan,
    pagebackref=true,
    pdftitle={ITSC2024},
    pdfpagemode=FullScreen,
    citecolor=green
    }
\def\BibTeX{{\rm B\kern-.05em{\sc i\kern-.025em b}\kern-.08em
    T\kern-.1667em\lower.7ex\hbox{E}\kern-.125emX}}
\begin{document}

\makeatletter
\let\@oldmaketitle\@maketitle
\renewcommand{\@maketitle}{\@oldmaketitle
  \centering
  \url{http://graphrelate3d.github.io}\\[8pt]
  \includegraphics[width=1.0\linewidth,trim={50, 170, 50, 93}]{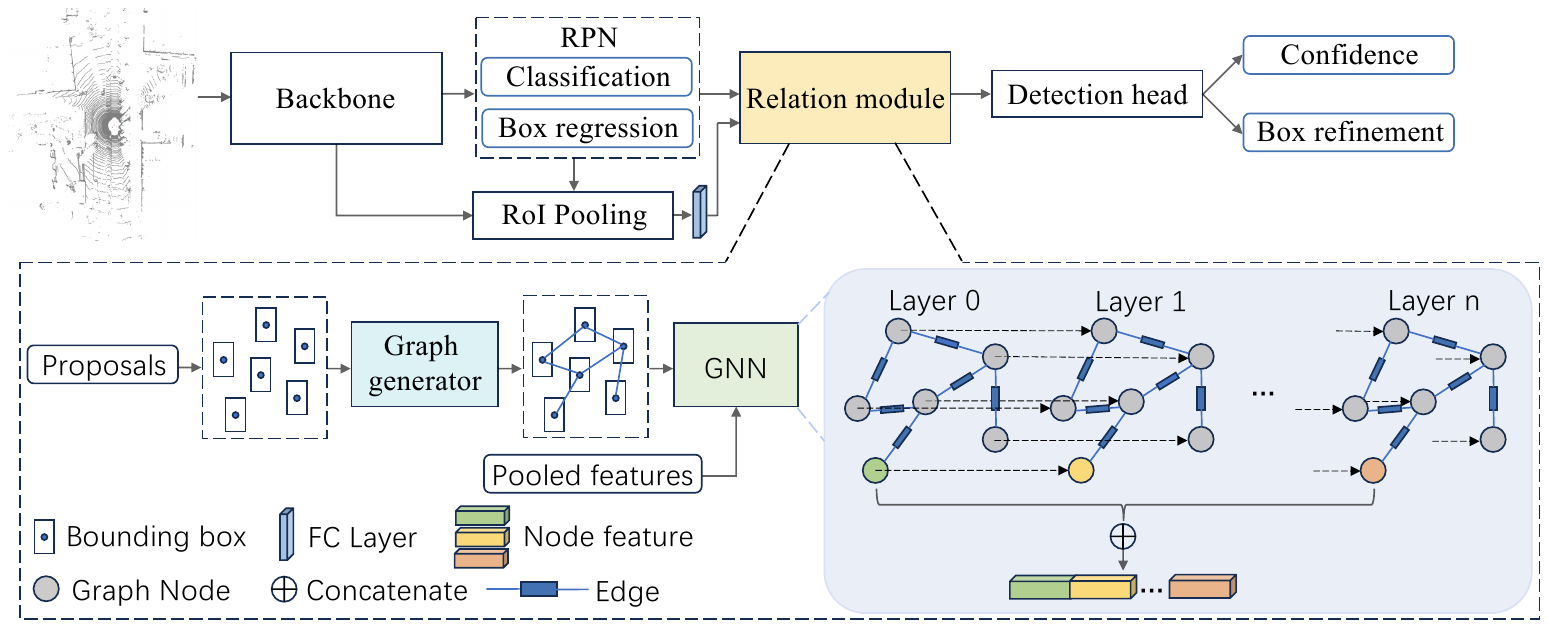}
  \captionof{figure}{Two-stage 3D object detection network extended with the proposed inter-object relation module (shown in \textcolor[RGB]{255,198,75}{yellow}). We implement our module after a detector's RPN and RoI pooling modules. The object relation module consists of a Graph Generator and a GNN. First, the Graph Generator conducts inter-object relation graphs based on the centers of proposal boxes. After that, the GNN takes the proposal features and box information to calculate and iteratively update the node features. Ultimately, the node features of the same node obtained by different layers are concatenated, which are the input of the following detection head.}
  \label{fig:pipeline}
  }  
\makeatother


\title{\LARGE \bf
GraphRelate3D: Context-Dependent 3D Object Detection with Inter-Object Relationship Graphs
}

\author{Mingyu Liu$^{\dagger}$~\orcidMingyu \qquad 
Ekim Yurtsever~\orcidEkim \qquad
Marc Brede~\orcidMarc \qquad
Jun Meng~\orcidJun \qquad
Walter Zimmer~\orcidWalter \\
Xingcheng Zhou~\orcidXingcheng \qquad
Bare Luka Zagar~\orcidLuka \qquad
Yuning Cui~\orcidYuning \qquad
Alois C. Knoll~\orcidKnoll
\thanks{M. Liu, M. Brede, J. Meng, W. Zimmer, X. Zhou, BL. Zagar, Y. Cui, and AC. Knoll are with the Chair of Robotics, Artiﬁcial Intelligence and Real-Time Systems, Technical University of Munich, 85748 Garching bei München, Germany}
\thanks{E. Yurtsever is with the College of Engineering, Center for Automotive Research, The Ohio State University, Columbus, OH 43212, USA}
\thanks{$^{\dagger}$ Corresponding author (mingyu.liu@tum.de)}
}

\maketitle
\setcounter{figure}{1}

\begin{abstract}
Accurate and effective 3D object detection is critical for ensuring the driving safety of autonomous vehicles. Recently, state-of-the-art two-stage 3D object detectors have exhibited promising performance. However, these methods refine proposals individually, ignoring the rich contextual information in the object relationships between the neighbor proposals. In this study, we introduce an object relation module, consisting of a graph generator and a graph neural network (GNN), to learn the spatial information from certain patterns to improve 3D object detection. Specifically, we create an inter-object relationship graph based on proposals in a frame via the graph generator to connect each proposal with its neighbor proposals. Afterward, the GNN module extracts edge features from the generated graph and iteratively refines proposal features with the captured edge features. Ultimately, we leverage the refined features as input to the detection head to obtain detection results. Our approach improves upon the baseline PV-RCNN on the KITTI validation set for the car class across easy, moderate, and hard difficulty levels by 0.82\%, 0.74\%, and 0.58\%, respectively. Additionally, our method outperforms the baseline by more than 1\% under the moderate and hard levels BEV AP on the test server. 
\end{abstract}


\section{Introduction}
\label{introduction}

Autonomous driving aims to reduce accidents and improve transportation efficiency by making the vehicles precisely perceive and intelligently interact with other agents in the surrounding environment~\cite{zhou2023vision}. Hence, an effective and robust perception system is pivotal for the safety of autonomous vehicles. 3D object detection is one of the most important tasks for ensuring a reliable perception system, which usually leverages the LiDAR point clouds as input to predict the category, bounding box size, and localization of objects. 

Prior works like PointPillars~\cite{lang2019pointpillars} divided input point clouds into several pillars and used feature extractors to obtain pillar features for detection. PV-RCNN~\cite{shi2020pv} and PartA$^{2}$~\cite{shi2019part} proposed two-stage frameworks for more accurate detection. To alleviate the data hunger challenge, SSL~\cite{erccelik20223d} and GCC-3D~\cite{liang2021exploring} explored self-supervised 3D object detection approaches. Moreover, VirConv~\cite{wu2023virtual} improved object detection accuracy by leveraging multi-modal information. However, the rich geometrical features of scenes are not utilized by these approaches.

In contrast, \cite{wang2021object, qian2022badet, schinagl2023gace, wu2022ret3d} introduced graphs to extract geometric information within scenarios to improve detection performance. Object DGCNN~\cite{wang2021object} represented the process of 3D object detection as the message passing through dynamically constructing and updating a graph. BADet~\cite{qian2022badet} constructed a local neighbor graph to model the local boundary correlations of an object. While these methods utilize graphs to harness geometric information effectively, they are limited to applying graphs either solely to high-dimensional features or to proposals of a single object.


\begin{figure}[t]
    \centering
    \begin{subfigure}{0.46\linewidth}
        \centering
        \includegraphics[width=\textwidth]{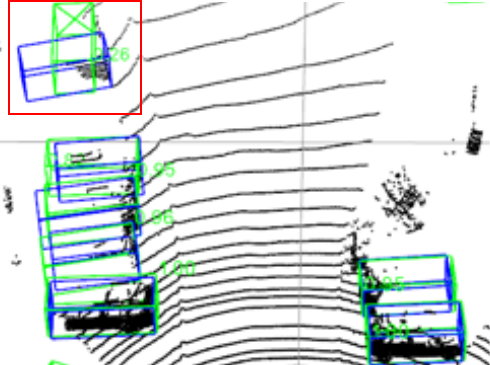}
        \caption{PV-RCNN}
    \end{subfigure}
    \hfill
    \begin{subfigure}{0.46\linewidth}
        \centering
        \includegraphics[width=\textwidth]{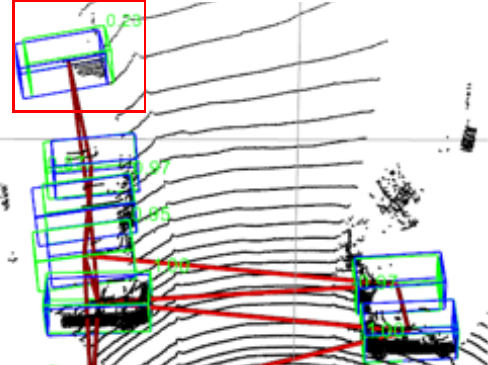}
        \caption{PV-RCNN Relation}
    \end{subfigure}
    \caption{Comparison between PV-RCNN~\cite{shi2020pv} (left) and our PV-RCNN with an object relation module (right). \textcolor{blue}{Blue} and \textcolor{green}{green} bounding boxes represent the ground truth and predictions, respectively. The \textcolor[rgb]{0.5451, 0, 0}{dark red} lines illustrate the relationship graph. As shown in the \textcolor{red}{red} box, our framework improves the bounding box rotation accuracy compared with the baseline PV-RCNN because of the information provided by similar patterns of the parked cars.}
    \label{fig:figure1}
\end{figure}

We agree that generating object relations is beneficial in extending the perception fields and improving detection accuracy under occlusion and distant objects. Moreover, certain patterns, such as parallel parking on narrow streets and interaction in multi-lane driving, can be explored to refine the predicted direction of objects (as shown in Fig.~\ref{fig:figure1}).
Hence, we introduce a novel 3D object detection module with inter-object relationship graphs to exploit the object patterns and geometrical relation, fully utilizing both spatial information and object features. We focus on improving the precision and robustness of two-stage object detectors by implementing our module on the proposals generated from the detectors. Our module can be easily implemented in various two-stage detectors. Specifically, we conduct an inter-object relationship graph within each point cloud frame. Additionally, our approach extracts object relation features from the boxes and features of proposals using a Graph Neural Network (GNN)~\cite{scarselli2008graph}. The box information implicitly includes object patterns. By leveraging the box information, each object's position, rotation angle, and size are incorporated into the graph and shared between neighbor nodes. Afterward, the object relation features are combined with the proposal features to provide local object features and global geometrical information. The integrated features are further as input to the detection head to obtain the final high-quality detection results.

Our main contributions are shown as follows:
\begin{itemize}
    \item We introduce a novel context-dependent 3D object-detection module with inter-object relationship graphs for intelligent vehicles. The proposed module learns object relationships and updates object features within the driving scenes by a GNN, which benefits addressing occlusions and detecting distant objects.
    \item We explore the information of proposal boxes to distinguish and leverage movement patterns within a scenario, such as the same rotation angle of parallel parking, to refine the directions of the predicted bounding boxes.
    \item Experiments on the real-world KITTI dataset verify the effectiveness of our proposed approach. In the official test server, our module outperforms the baseline PV-RCNN model by 1.13\% and 1.22\% for car class on moderate and hard difficulty levels, respectively.
\end{itemize}

\section{Related Works}
\label{related_work}

\subsection{3D Object Detection}
\label{3d_od}
LiDAR point clouds provides rich spatial information for 3D object detection. Usually, 3D object detectors extract features from point clouds with approaches, such as voxelization~\cite{lang2019pointpillars}, graph neural netork~\cite{shi2020point}, range view~\cite{bai2024rangeperception}, or bird’s eye view (BEV)~\cite{ku2018joint}. PointPillars~\cite{lang2019pointpillars}, as a pioneering work, splits points into a set of pillars, achieving efficient inference performance, while its detection accuracy is limited. MSF~\cite{he2023msf} explores temporal information from sequential frames for 3D object detection. Inspired by 2D object detection, some works employ two-stage detectors to predict more precise and robust detection results than one-stage networks. PV-RCNN~\cite{shi2020pv} proposes a two-stage framework that utilizes both the voxel and point features and a multi-scale RoI feature abstraction layer for prediction refinement. PartA$^{2}$~\cite{shi2019part} introduces a detector consisting of part-aware and part-aggregation stages. Although these approaches improve detection, their performance is limited under occlusion conditions.

\subsection{GNNs in LiDAR point clouds} 
\label{graph_neural_network}
Graph neural networks (GNNs) capture graph features via message passing between nodes within the graph~\cite{zhou2020graph}, making it suitable to represent complex relationships between vertices. Several works represent point clouds as graphs and extract their features via GNNs. Bi et al.~\cite{bi2019graph} leverages a GNN for point cloud-based object classification. DGCNN~\cite{wang2019dynamic} introduces a framework that dynamically computes and updates graph features in each layer for classification and segmentation. 
These methods exploit applying GNNs to extract local point features, verifying the capability of GNNs in processing complex graphs. 

\subsection{Graph-based 3D Object Detection}
\label{object_relation}
Motivated by the excellent performance of GNNs, several works explore graphs for 3D object detection in autonomous driving.
Point-GNN~\cite{shi2020point} generates graphs based on irregular points and utilizes a GNN to obtain features, showing promising detection performance. BADet~\cite{qian2022badet} introduces a boundary-aware 3D detector, which generates a local graph for each object based on its proposals. Besides only utilizing a single frame, Ret3D~\cite{wu2022ret3d} utilizes a graph and a transformer to refine detection in sequential frames. Object DGCNN~\cite{wang2021object} incorporates features from the Bird's-Eye View (BEV) grid and further incorporates several queries to refine detection. GACE~\cite{schinagl2023gace} enhances detection confidence by leveraging point features within predicted bounding boxes and those from adjacent boxes in the vicinity. This optimization is achieved through a multilayer perceptron (MLP). In comparison to the previous works, our work generates an object-relation graph using proposals of various objects within a single frame. A GNN is used to learn relation information from the differences in bounding boxes and proposal features. Our approach efficiently leverages geometric information and movement patterns involved in object relations to enhance the 3D perception systems of intelligent vehicles.

\section{Methodology}\label{methodology}


This study introduces our novel inter-object relation module for improving 3D object detection in autonomous driving. Traditionally, two-stage 3D detectors~\cite{shi2020pv, shi2019part} process each proposal independently during the refinement stage, ignoring the rich context information included in the object relations. However, learning the inter-object relationship is beneficial for more effective and robust detection, especially for addressing occlusion and predicting object direction. To do so, we introduce an object relation module to enlarge the perception range and capture relation features by leveraging a GNN on the generated inter-object relationship graphs. The overview of our proposed method is illustrated in Fig.~\ref{fig:pipeline}. In the following, we explain the proposed approach in detail. 

\subsection{Preliminaries}
We focus on improving the performance of two-stage 3D object detectors. We denote the set of 3D proposals in a frame generated by the Region Proposal Network (RPN) in the first stage of a 3D detector as $P = \left\{\mathbf{p}_{i}|i=1,\dots,n\right\}$, where $n$ is the total number of predicted proposals in a frame. The proposals include the classification $C \in \mathbb{R}^{n \times 1}$ and coarse predicted bounding boxes $B \in \mathbb{R}^{n \times 7}$. Each bounding box consists of box size $(h, w, l)$, center $(x, y, z)$, and a heading angle $\theta$. After that, a Region of Interest (RoI) pooling module combined with a fully connected (FC) layer is used to create the fixed-size feature maps $F = \left\{\mathbf{f}_{i}\in \mathbb{R}^{1 \times d}|i=1,\dots,n\right\}$, where $d$ is the dimension of features. In the end, the feature maps are input to a set of layers to obtain the final predicted bounding boxes $B^{res} = \left\{\mathbf{b}^{res}_{i^{\prime}}|i^{\prime}=1, \dots, m\right\}$ and their confidence scores $C_{score} \in \mathbb{R}^{1 \times m}$, where $m$ is the number of predicted 3D bounding boxes. 


In our approach, we fully use graph information while preserving local object features by incorporating our proposed inter-object relation module after the FC layer. This proposed module generates refined features $F^{\prime} \in \mathbb{R}^{n \times d^{\prime}}$ from the RoI feature maps $F$ and the coarse bounding boxes $B$. 

\subsection{Inter-Object Relationship Graphs}
\begin{figure}[t]
    \centering
    \begin{subfigure}{0.38\linewidth}
        \centering
        \includegraphics[width=\textwidth]{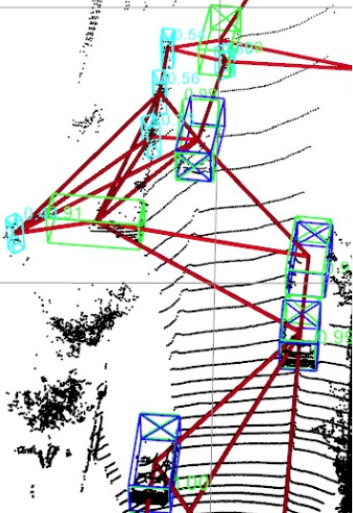}
        \caption{KNN (k=16)}
    \end{subfigure}
    \hspace{20pt}
    \begin{subfigure}{0.38\linewidth}
        \centering
        \includegraphics[width=\textwidth]{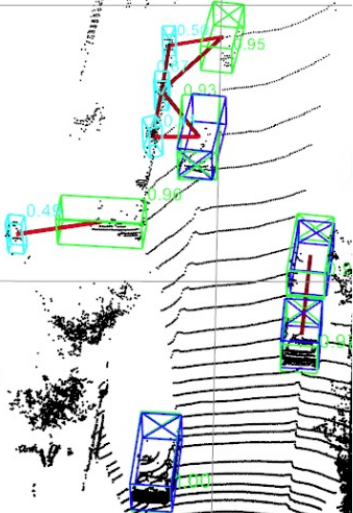}
        \caption{Radius (r=6m)}
    \end{subfigure}
    \caption{Example of generated graphs on proposals from the first stage of PV-RCNN~\cite{shi2020pv}. The left graph was generated based on KNN (K=16), leading every proposal to connect to its 16 nearest neighbors. The right diagram was generated via a radius graph with a threshold of six meters. Ground truth is shown in \textcolor{blue}{blue}. \textcolor[rgb]{0.5451, 0, 0}{Dark red} lines illustrate the graph edges. \textcolor{green}{Green} and \textcolor[rgb]{0.0902, 0.9922, 0.9922}{light blue} represent the predicted cars and pedestrians.}
    \label{fig:graph_example}
\end{figure}

We create an inter-object relationship graph based on the predicted proposals $P$ in a point cloud. Creating a fully connected graph among all proposals requires the runtime complexity to $O(n^{2})$, where $n$ is the number of proposals. On the other hand, connecting all proposals can make a large receptive field while causing noise and introducing unrelated information from distant objects. Therefore, we consider establishing the graph based on box centroid $(x,y,z)$ within a predefined range. Specifically, we utilize two strategies to construct a graph: k nearest neighbor (KNN), connecting a fixed number of proposals with each other, or a radius graph, connecting all proposals within a certain distance. After graph generation, each proposal connects to its neighbors, as shown in Fig.~\ref{fig:graph_example}.

We follow~\cite{wu2020comprehensive} to formulae a graph as $G=(V, E)$, where $V$ is the set of nodes and $E$ is the set of edges. An edge pointing from the node $i$ to $j$ can be illustrated as 
$(i, j) \in E$, where $j \in \mathcal{N}(i)$ is one of the neighbor nodes of node $i$. Since we utilize directed graphs for message passing between connected nodes, edge $(i, j)$ and edge $(j, i)$ need to be distinguished.

\begin{figure}[htb]
    \centering
    \includegraphics[width=0.48\textwidth, trim={200, 190, 160, 200}]{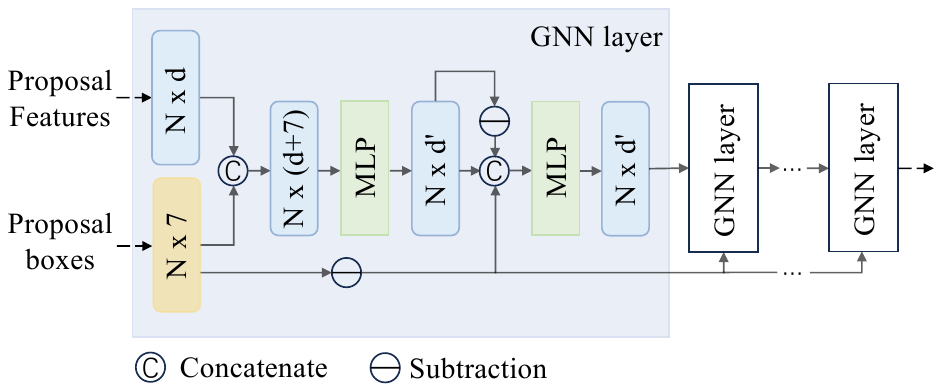}
    \caption{Illustration of Graph Neural Network architecture. We utilize the proposal features and boxes as input to initialize the node features of the inter-object relationship graph. After a set of GNN layers, the object relation module outputs the refined feature of each proposal.}
    \label{fig:gnn_layer}
\end{figure}

Generally, a process of updating node features via a graph neural network can be demonstrated as:

\begin{equation}
    \mathbf{v}^{l+1}_{i} = f(\mathbf{e}^{l}_{ij}, \mathbf{v}^{l}_{i})
    \label{equation1}
\end{equation}
where $l \in L$ is the $l$th layer of the GNN with $L$ layers in total, $\mathbf{v}_{i}$ is the feature of the $i$th node, $\mathbf{e}_{ij}$ is the feature of edge $(i, j)$, and $f(.)$ is the operation applying edge features to update the node features. Usually, edge feature $\mathbf{e}^{l}_{ij}$ can be defined as $\mathbf{e}^{l}_{ij} = h^{l}(\mathbf{v}^{l}_{i}, \mathbf{v}^{l}_{j})$, where $h^{l}(.)$ is a function to capture the edge feature. 
 
We demonstrate the GNN module we used in our graph relation module in Fig.~\ref{fig:gnn_layer}. We initialize each node feature $\mathbf{v}^{0}_{i}$ by fusing the proposal feature obtained by the FC layer and the corresponding box information $\mathbf{b}_{i}$ with an MLP layer, as shown in Eq.~\ref{initial_node_features}. Therefore, each node feature compresses object and spatial features by fusing the proposal features with the proposal box information.  
\begin{equation}
    \mathbf{v}^{0}_{i} = MLP(\mathbf{f}_{i}, \mathbf{b}_{i})
    \label{initial_node_features}
\end{equation}

We first leverage the differences of node features to obtain the edge features. Encompassing features from neighbor nodes enlarge the receptive field, alleviating occlusion issues and improving the capability to perceive the surrounding environment.
Hence, Eq.~\ref{equation1} is re-written as:
\begin{equation}
    \mathbf{v}^{l+1}_{i} = g^{l}(h^{l}(\mathbf{v}^{l}_{j} - \mathbf{v}^{l}_{i}, \mathbf{v}^{l}_{i}))
    \label{updated_node_features}
\end{equation}

Additionally, we add the proposal box difference between the neighbor nodes into edge features to utilize the inter-object relations more effectively. The proposal box differences directly reflect the spatial relationships between neighbor objects, allowing the GNN to capture relative positioning, which helps understand the movement patterns of objects and their interaction with each other.

Specifically, we concatenate the proposal box differences and node feature differences to obtain the enhanced edge features. For Eq.~\ref{updated_node_features}, we conduct the function $h^{l}(.)$ as MLP to extract edge features, and $g^{l}(.)$ as max pooling to update the node feature $\mathbf{v}^{l}_{i}$ to $\mathbf{v}^{l+1}_{i}$ with the aggregating edge features. Hence, the final node feature updated by the $l+1$ layer can be represented as: 
\begin{equation}
   \mathbf{v}^{l+1}_{i} = \max_{\forall j \in \mathcal{N}(i)}(MLP(\mathbf{v}^{l}_{j} - \mathbf{v}^{l}_{i}, \mathbf{b}_{j} - \mathbf{b}_{i}, \mathbf{v}^{l}_{i}))
   \label{final_node_features}
\end{equation}

Since a node feature $v_{i}$ is iteratively updated through multiple GNN layers, it encodes different hidden features after each layer. Hence, to keep all the hidden features learned by the GNN, we combine all the features of the same node from different layers:
\begin{equation}
    \mathbf{f}^{\prime}_{i} = (\mathbf{f}^{0}_{i}, \mathbf{f}^{1}_{i}, \dots, \mathbf{f}^{L}_{i})
\end{equation}
Ultimately, we use the combined node features to input the following classification head to produce the refined object bounding boxes and predicted confidence score.

\section{Experiments and Results}
\label{excperiments_results}

\subsection{Dataset}
We utilize the KITTI 3D Object Detection dataset~\cite{geiger2012we}, a fundamental dataset widely used in autonomous driving~\cite{liu2024survey}, to train and evaluate our proposed inter-object relationship method. The KITTI dataset consists of 7,481 annotated LiDAR scans sampled from different driving scenes. We utilize the common train-val split with 3712 training and 3769 validation samples to train and evaluate the baseline and our model. 

\noindent\textbf{Evaluation Metric.} The KITTI 3D average precision (AP) and Bird's-Eye-View (BEV) AP metrics are used in our experiments. We leverage the IoU threshold to 0.7 for the car class and 0.5 for pedestrians and cyclists under three difficulty levels (easy, moderate, and hard). All AP values are calculated with 40 recall positions on the validation set and the official test server.

\subsection{Experimental Setup}
We conduct all experiments using the OpenPCDet\footnote{\color{red}{\url{https://github.com/open-mmlab/OpenPCDet}}} repository based on PyTorch with one NVIDIA GeForce 3090 (24GB) GPU. 

\noindent\textbf{Hyperparameters.} For all experiments, we set the batch size to 2, using the Adam OneCycle optimizer with an initial learning rate of 0.01. All models are trained for 80 epochs, and we report the best result of each experiment in the experiment tables. 

\noindent\textbf{Network Architecture.}  We use PV-RCNN~\cite{shi2020pv} and PartA$^{2}$~\cite{shi2019part} as the baseline models and implement our proposed object relation module in them. For both PV-RCNN and PartA$^{2}$, we keep using the same model architectures reported in \cite{shi2020pv} and \cite{shi2019part}, respectively. For the proposed module, we apply KNN (k = 16) as the graph generator for the main experiments and use a four-layer GNN to extract features. To keep the feature dimension alignment, we set the input and output of the GNN module to 256, the same as the dimensions of the pooled features and the input features of the detection head of the two-stage 3D detectors.

\noindent\textbf{Training Loss.}
Our proposed object relation module does not introduce external loss calculation, making our module easily adapted to any other two-stage detectors. Therefore, in our experiments, we follow the default loss function of PV-RCNN and PartA$^{2}$ to conduct experiments.
\begin{figure*}[htb]
    \centering
    \begin{subfigure}{0.32\textwidth}
    \centering
        \includegraphics[width=\linewidth, trim={0, 0, 30, 0}]{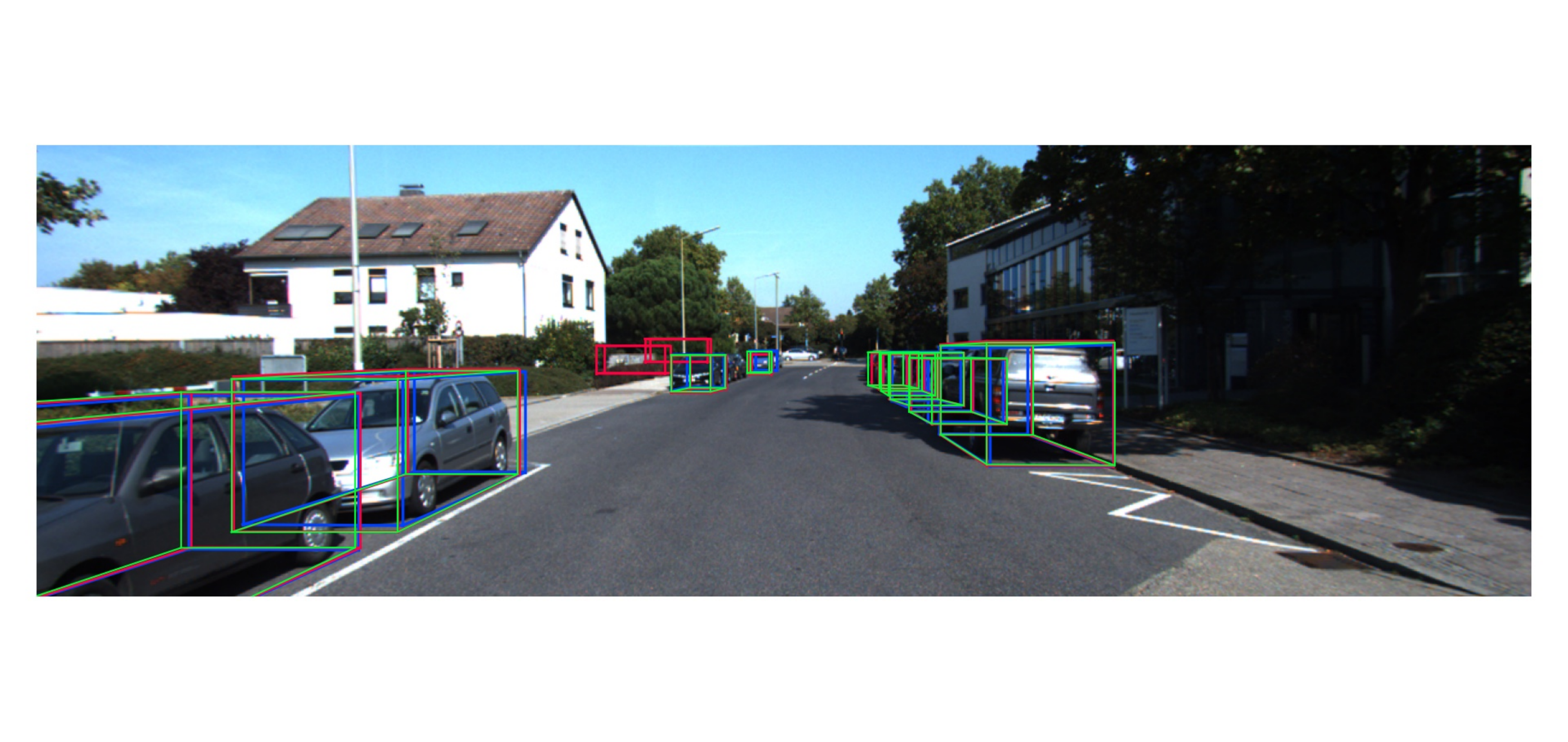}
        \includegraphics[width=0.66\linewidth, trim={0, 0, 0, 0}]{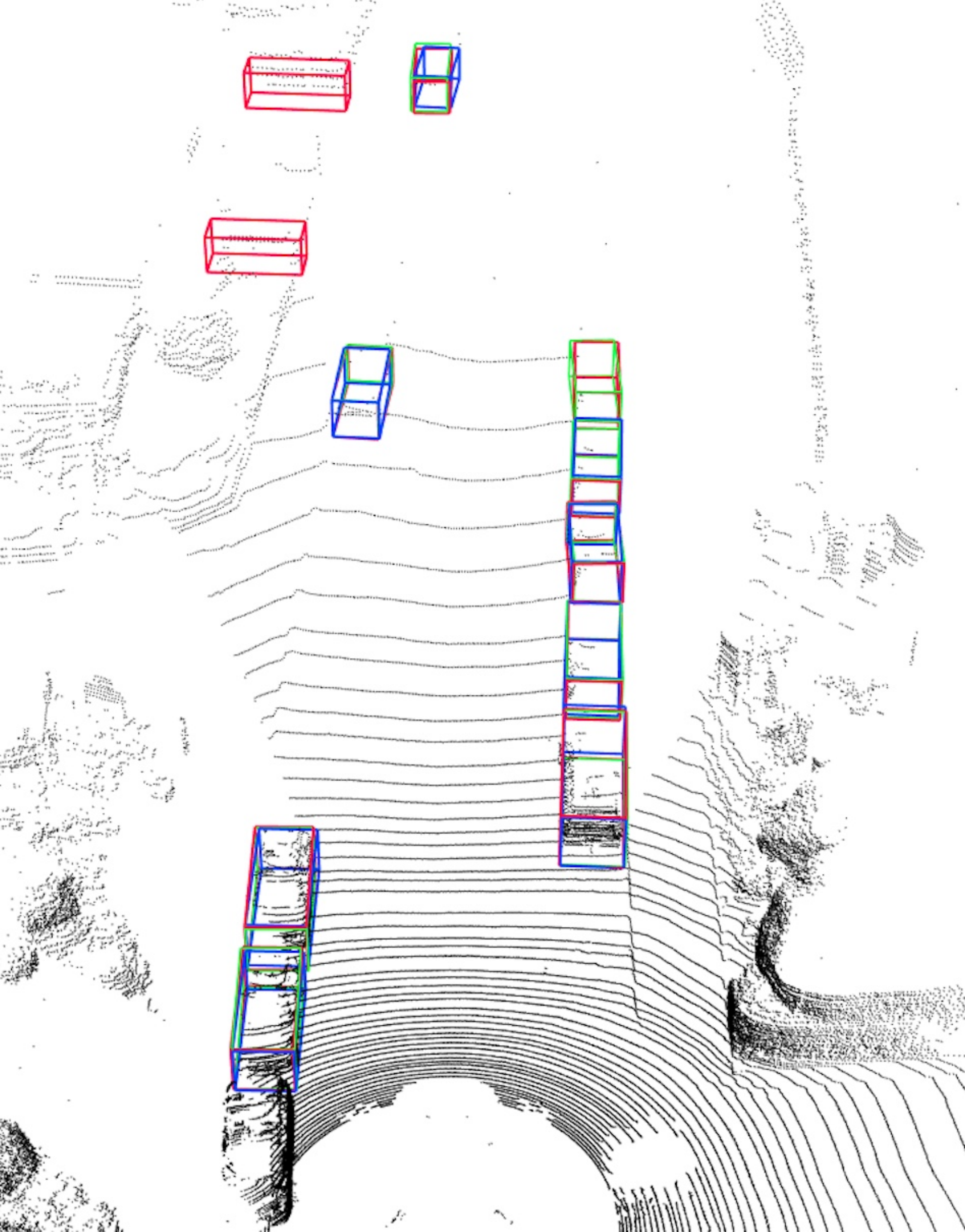}
        \caption{frame 159}
    \end{subfigure}
    \hfill
    \begin{subfigure}{0.32\textwidth}
    \centering
        \includegraphics[width=\linewidth, trim={30, 0, 30, 0}]{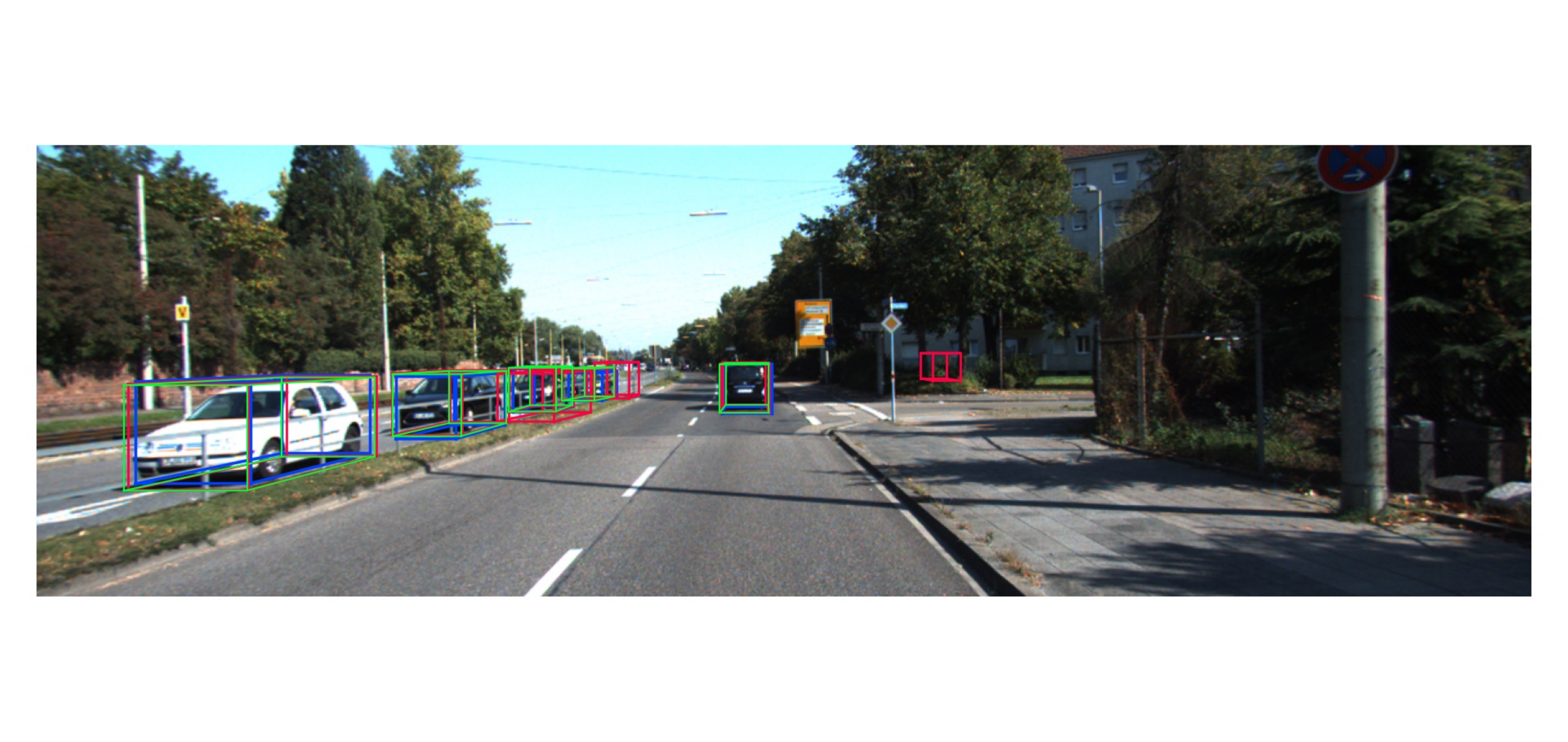}
        \includegraphics[width=0.718\linewidth, trim={0, 0, 0, 0}]{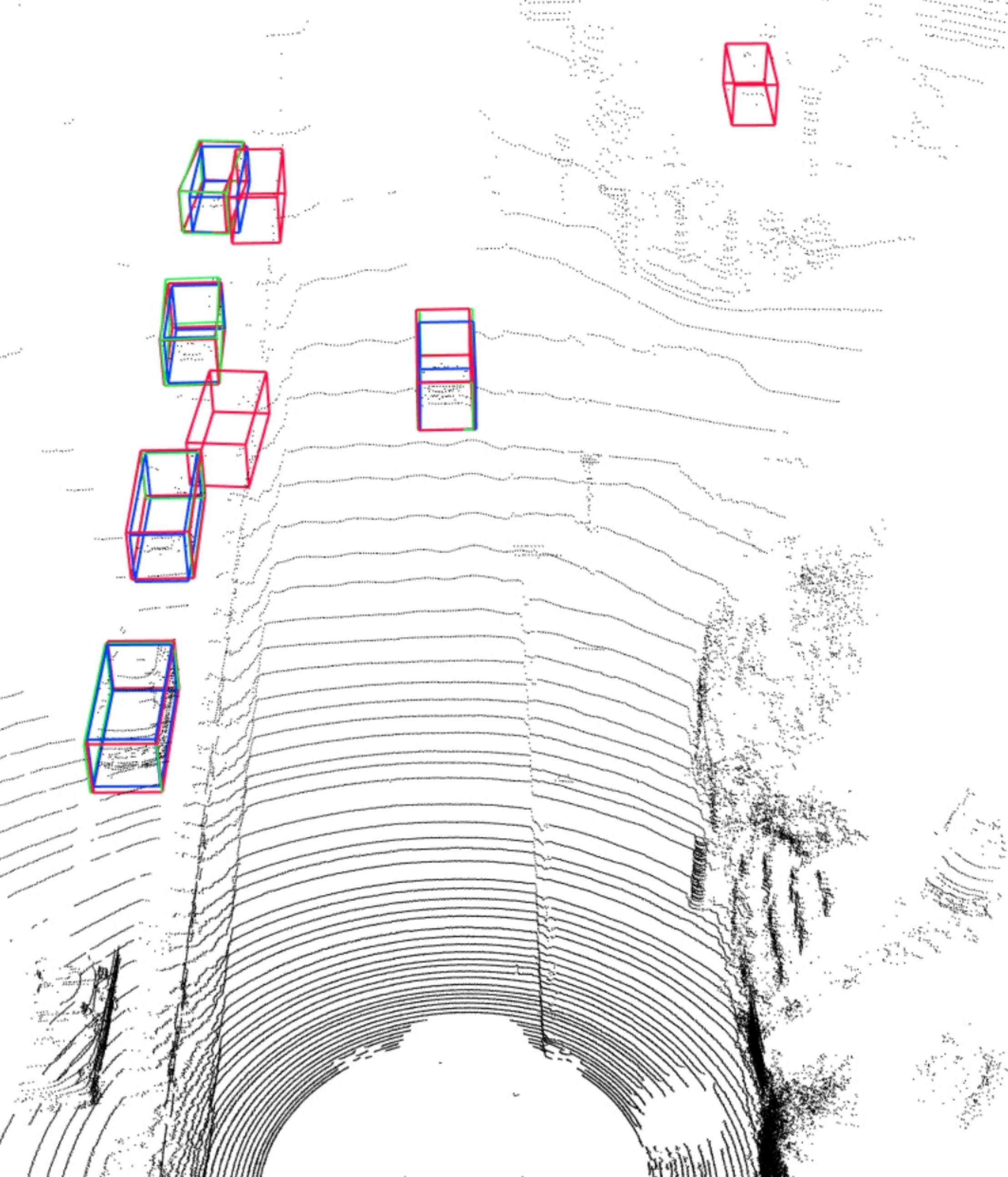}
        \caption{frame 216}
    \end{subfigure}
    \hfill
    \begin{subfigure}{0.32\textwidth}
    \centering
        \includegraphics[width=\linewidth, trim={30, 0, 0, 0}]{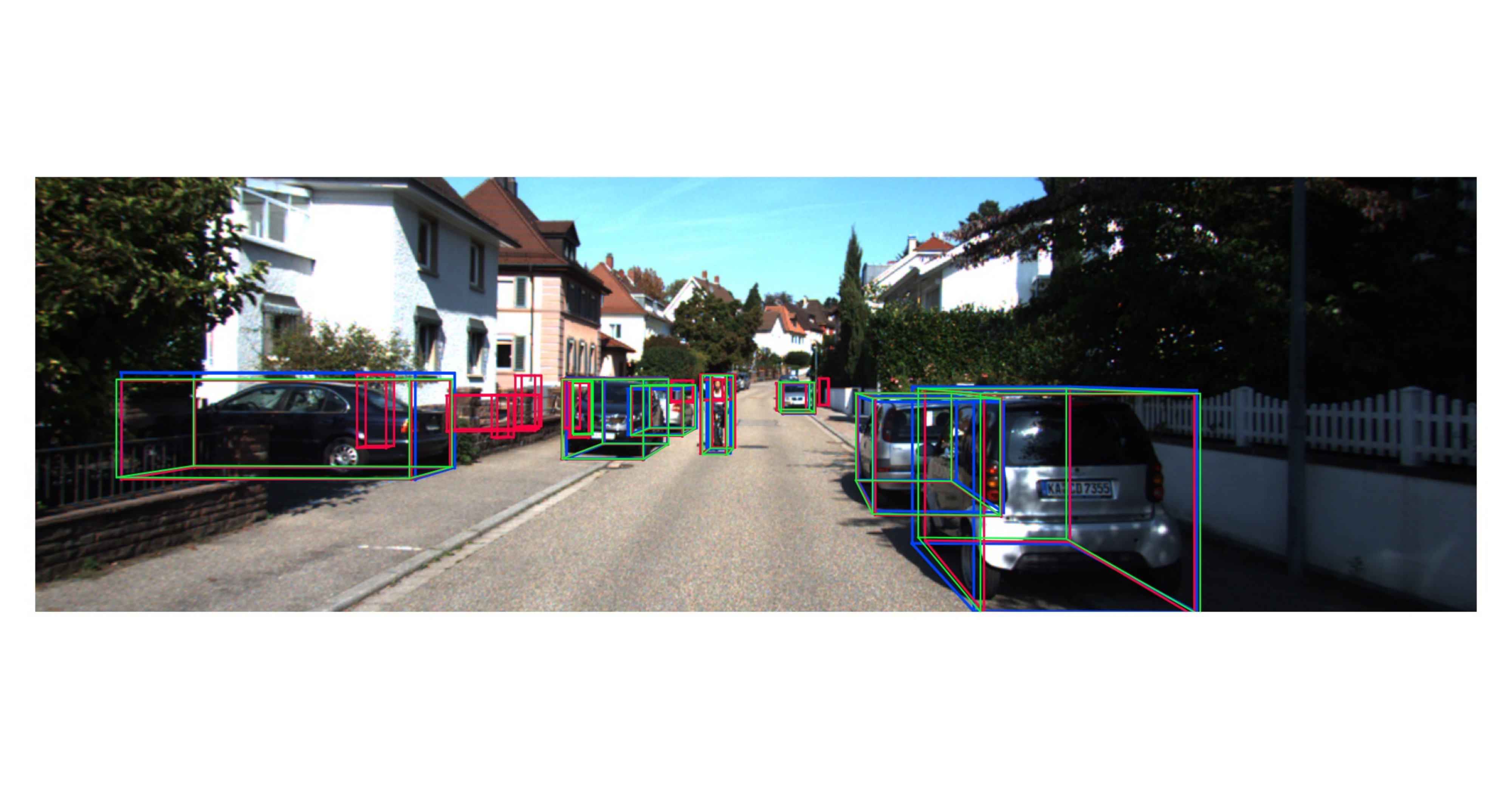}
        \includegraphics[width=0.56\linewidth, trim={0, 0, 0, 0}]{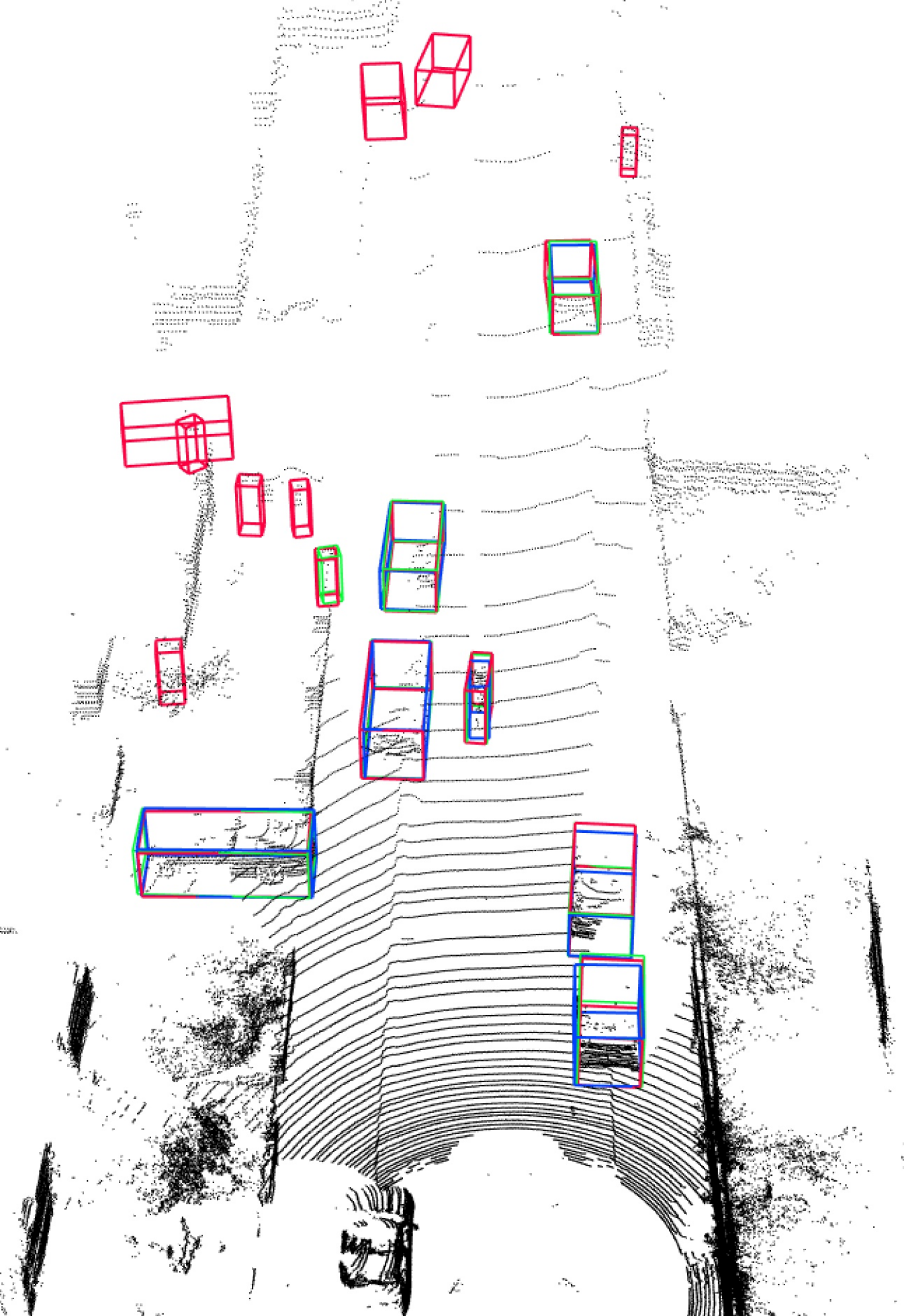}
        \caption{frame 335}
    \end{subfigure}
    
    
    \caption{Qualitative results on KITTI \textbf{validation set}. We demonstrate detection results from three scenarios. \textbf{Ground truths}, the results of \textbf{baseline}, and \textbf{ours} are shown in \textcolor{blue}{blue}, \textcolor{red}{red}, and \textcolor{green}{green}, respectively. By learning similar movement patterns and extracting relation information, our approach significantly reduces false positives and improves predicted rotation accuracy. Zoom in for more details.}
    \label{fig:car_qualitative_res}
\end{figure*}

\noindent\textbf{Data Augmentation.} We utilize four widely adopted data augmentation strategies in our experiments: 1) Ground truth sampling~\cite{yan2018second}, randomly selecting several ground truths from other scenes and adding them into the current frame. 2) Random flipping along the x-axis. 3) Random rotation within the angle range $[-\frac{\pi}{4}, \frac{\pi}{4}]$ around the z axis. 4) Random scaling with a scaling factor within the range $[0.95, 1.05]$.

\subsection{Results and Discussion}

\begin{table*}[htb]
    \centering
    \caption{Comparison with the baseline models (PV-RCNN~\cite{shi2020pv} and PartA$^{2}$~\cite{shi2019part}) with our proposed object relation module on the KITTI~\cite{geiger2012we} \textbf{validation set} in terms of 3D AP$_{R40}$ and BEV AP$_{R40}$. For the car and cyclist categories, the IoU thresholds are 0.7 and 0.5, respectively.}
    \resizebox{\linewidth}{!}{
    \begin{tabular}{cccccccccccccc}
    \toprule[1pt]
    \multirow{2}{*}{Method} & \multicolumn{3}{c}{Car 3D AP (\%)~$\uparrow$} & \multicolumn{3}{c}{Car BEV AP (\%)~$\uparrow$} & \multicolumn{3}{c}{Cyclist 3D AP (\%)~$\uparrow$} & \multicolumn{3}{c}{Cyclist BEV AP( \%)~$\uparrow$} \\
     ~ & Easy & Moderate & Hard & Easy & Moderate & Hard & Easy & Moderate & Hard & Easy & Moderate & Hard \\
    \midrule
    PartA$^{2}$ & \textbf{91.84} & 82.53 & 80.17 & \textbf{94.05} & 88.33 & 88.04 & 88.24 & \textbf{71.12} & 66.76 & 89.14 & 72.87 & 69.59 \\
    PartA$^{2}$ Relation (ours) & 91.72 & \textbf{82.59} & \textbf{80.43} & 93.08 & \textbf{88.90} & \textbf{88.72} & \textbf{90.78} & 70.93 & \textbf{67.60} & \textbf{93.20} & \textbf{75.08} & \textbf{71.66} \\
    \rowcolor{gray!20}
    Improvement & \textcolor{red}{-0.12} & \textcolor[RGB]{50,205,50}{+0.06} & \textcolor[RGB]{50,205,50}{+0.26} & \textcolor{red}{-0.97} & \textcolor[RGB]{50,205,50}{+0.57} & \textcolor[RGB]{50,205,50}{+0.68} & \textcolor[RGB]{50,205,50}{+2.54} & \textcolor{red}{-0.19} & \textcolor[RGB]{50,205,50}{+0.84} & \textcolor[RGB]{50,205,50}{+4.06} & \textcolor[RGB]{50,205,50}{+2.21} & \textcolor[RGB]{50,205,50}{+2.07} \\
    \midrule
    PV-RCNN & 91.87 & 84.53 & 82.41 & 94.58 & 91 & 88.51 & 89.95 & 70.81 & 66.37 & 91.26 & 73.94 & 69.53 \\
    PV-RCNN Relation (ours) & \textbf{92.45} & \textbf{85.36} & \textbf{83} & \textbf{95.54} & \textbf{91.57} & \textbf{89.26} & \textbf{91.52} & \textbf{71.13} & \textbf{66.59} & \textbf{93.5} & \textbf{74.55} & \textbf{69.81} \\
    \rowcolor{gray!20}
    Improvement & \textcolor[RGB]{50,205,50}{+0.58} & \textcolor[RGB]{50,205,50}{+0.83} & \textcolor[RGB]{50,205,50}{+0.59} & \textcolor[RGB]{50,205,50}{+0.96} & \textcolor[RGB]{50,205,50}{+0.57} & \textcolor[RGB]{50,205,50}{+0.75} & \textcolor[RGB]{50,205,50}{+1.57} & \textcolor[RGB]{50,205,50}{+0.32} & \textcolor[RGB]{50,205,50}{+0.22} & \textcolor[RGB]{50,205,50}{+2.24} & \textcolor[RGB]{50,205,50}{+0.61} & \textcolor[RGB]{50,205,50}{+0.28} \\
     \bottomrule[1pt]  
    \end{tabular}
    }
    \label{tab:res_all_classes_kitti_val}
\end{table*}
We conduct several experiments to show the efficacy of our inter-object relation module in improving two-stage 3D object detection.
\noindent\textbf{Results on KITTI Validation Set.} We first train PV-RCNN~\cite{shi2020pv} baseline and PV-RCNN with our relation module on the KITTI training set and evaluate the prediction results on the KITTI validation set. In Tab.~\ref{tab:res_all_classes_kitti_val}, we demonstrate the results of the baseline PV-RCNN and ours. Our PV-RCNN relation network outperforms the baseline with accuracies of 92.45\%, 85.36\%, and 83\% in terms of the 3D AP for car class under easy, moderate, and hard difficulties. Meanwhile, our model performs better on the BEV AP than the baseline PV-RCNN with clear improvements (easy: 0.96\%, moderate: 0.57\%, and hard: 0.75\%). The improvements are attributed to the contextual patterns learned by our graph relational module, which effectively mitigates occlusion issues, boosting detection accuracy. Additionally, we also report the experiment results based on PartA$^{2}$~\cite{shi2019part} in Tab.~\ref{tab:res_all_classes_kitti_val}. In general, our object relation module in PartA$^{2}$ exhibits a similar capability as in PV-RCNN. Although there is a slight performance decrease in terms of the easy-difficulty car class, our proposed module increases the BEV detection accuracy of the moderate and hard difficulties by 0.57\% and 0.68\%. 

Moreover, we note that for PV-RCNN and PartA$^{2}$, the detection performance of cyclists is promisingly optimized by the proposed object relation module. For example, PartA$^{2}$ relation surpasses the baseline with 4.06\%, 2.21\%, and 2.07\% for BEV AP under the easy, moderate, and hard difficulty levels. In most cases, only a few cyclists are typically present in a scene (Fig.~\ref{fig:car_qualitative_res} (c)), which can be easily obscured by vehicles or pedestrians. By establishing object relationships with nearby proposals, a broader receptive field and object relation are created. It enables the network to effectively capture the distinctive movement patterns of cyclists, thereby enhancing detection accuracy.

\begin{table}[htb]
    \centering
    \caption{Comparison with other state-of-the-art methods, the baseline models (PV-RCNN~\cite{shi2020pv} and PartA$^{2}$~\cite{shi2019part}), and ours on the KITTI~\cite{geiger2012we} \textbf{validation set} for car class using 3D AP$_{R40}$ and BEV AP$_{R40}$ with IoU threshold 0.7.}
    \label{tab:only_car_kitti_val}
    \resizebox{\linewidth}{!}{
    \begin{tabular}{cccccccc}
    \toprule[1pt]
        \multirow{2}{*}{Method} & \multicolumn{3}{c}{Car - 3D AP (\%)~$\uparrow$} & \multicolumn{3}{c}{Car - BEV AP (\%)~$\uparrow$} \\
        ~ & Easy & Mod. & Hard & Easy & Mod. & Hard \\
        \midrule
        SECOND~\cite{yan2018second} & 87.43 & 76.48 & 69.10 & 89.96 & 87.07 & 79.66 \\ 
        CIA-SSD~\cite{zheng2021cia} & 90.04 & 79.81 & 78.80 & - & - & - \\
        SSL Point-GNN~\cite{erccelik20223d} & 91.43 & 82.85 & 80.12 & 93.55 & 89.79 & 87.23 \\
        PointRCNN~\cite{shi2019pointrcnn} & 88.88 & 78.63 & 77.38 & - & - & - \\
        \midrule
        PartA$^{2}$ & 92.28 & 82.70 & 80.41 & 93.33 & \textbf{89.69} & 88.40 \\
        PartA$^{2}$ Relation (ours) & \textbf{92.53} & \textbf{83.15} & \textbf{80.88} & \textbf{95.89} & 89.45 & \textbf{89.19} \\
        \rowcolor{gray!20}
        Improvement & \textcolor[RGB]{50,205,50}{+0.25} & \textcolor[RGB]{50,205,50}{+0.45} & \textcolor[RGB]{50,205,50}{+0.47} & \textcolor[RGB]{50,205,50}{+2.56} & \textcolor{red}{-0.24} & \textcolor[RGB]{50,205,50}{+0.79} \\
        \midrule
        PV-RCNN & 91.91 & 84.78 & 82.63 & 93.17 & 90.72 & 88.73 \\
        PV-RCNN Relation (ours) & \textbf{92.73} & \textbf{85.52} & \textbf{83.21} & \textbf{95.48} & \textbf{91.25} & \textbf{89.09} \\
        \rowcolor{gray!20}
        Improvement & \textcolor[RGB]{50,205,50}{+0.82} & \textcolor[RGB]{50,205,50}{+0.74} & \textcolor[RGB]{50,205,50}{+0.58} & \textcolor[RGB]{50,205,50}{+2.31} & \textcolor[RGB]{50,205,50}{+0.53} & \textcolor[RGB]{50,205,50}{+0.36} \\
    \bottomrule[1pt]
    \end{tabular}
    }
\end{table}
\noindent\textbf{Results on the Car Class.} To further validate the effectiveness of the proposed object relation module, we evaluate our method on the KITTI data using only car class for training and evaluation. We report the experiment results in Tab.~\ref{tab:only_car_kitti_val}. 
Compared to the baseline PV-RCNN, our relation module increases the 3D AP by 0.82\%, 0.74\%, and 0.58\% on easy, moderate, and hard difficulty levels, respectively. For the BEV AP, we achieve improvements of 2.31\%, 0.53\%, and 0.36\%. Furthermore, our object relation module also improves the detection performance of PartA$^{2}$ by 0.47\% and 0.79\% w.r.t the 3D AP and BEV AP on the hard level. The improvements provided by our method are critical for the driving safety of autonomous vehicles. We exhibit the qualitative results in Fig.~\ref{fig:car_qualitative_res}. The selected scenarios include cars parking on the roadside (Fig.~\ref{fig:car_qualitative_res} (a)) or driving parallel (Fig.~\ref{fig:car_qualitative_res} (b)), which are suitable for verifying our hypothesis of exploiting certain patterns to improve object detection. Our object relation module successfully reduces the two false positives shown in the top right of Fig.~\ref{fig:car_qualitative_res} (a), which parks horizontally, while the true positives park along the driving direction. Fig.~\ref{fig:car_qualitative_res} (b) further verifies the effectiveness of our method.

\noindent\textbf{Results on the KITTI Test Set.} To further verify the capability of our proposed object relation module, we evaluate our model on the official KITTI test server and compare ours with other algorithms from the leaderboard (see Tab.~\ref{tab:res_all_classes_kitti_test}). Due to the different experiment settings, we also report the performance of the baseline PV-RCNN trained by ourselves to ensure a fair comparison. Our PV-RCNN relation network performs better than the baseline with significant improvements. Especially for the BEV metric, more than 1\% accuracy improvements on the moderate and hard levels reflect the reliable ability of our method in challenging autonomous driving environments.

\begin{table}[htb]
    \centering
    \caption{Comparison with the state-of-the-art methods on the KITTI~\cite{geiger2012we} \textbf{test set} for car class using 3D AP$_{R40}$ and BEV AP$_{R40}$ with IoU threshold 0.7.}
    \resizebox{\linewidth}{!}{
    \begin{tabular}{cccccccc}
    \toprule[1pt]
         \multirow{2}{*}{Method} & \multicolumn{3}{c}{Car - 3D AP (\%)~$\uparrow$} & \multicolumn{3}{c}{Car - BEV AP (\%)~$\uparrow$} \\
         ~ & Easy & Moderate & Hard & Easy & Moderate & Hard \\
         \midrule
         SECOND~\cite{yan2018second} & 84.65 & 75.96 & 68.71 & 91.81 & 86.37 & 81.04 \\ 
         PointPillars~\cite{lang2019pointpillars} & 82.58 & 74.31 & 68.99 & 90.07 & 86.56 & 82.81 \\
         Fast Point R-CNN~\cite{chen2019fast} & 85.29 & 77.40 & 70.24 & 90.87 & 87.84 & 80.52 \\
         PartA$^2$~\cite{shi2019part} & 87.81 & 78.49 & 73.51 & 91.70 & 87.79 & 84.61 \\
         RangeDet~\cite{fan2021rangedet} & 85.41 & 77.36 & 72.60 & 90.93 & 87.67 & 82.92 \\
         SSL PointPillars~\cite{erccelik20223d} & 82.54 & 72.99 & 67.54 & 88.92 & 85.73 & 80.33 \\
         \midrule
         PV-RCNN & 87.20 & 78.76 & 75.50 & 91.97 & 87.90 & 85.02 \\
         PV-RCNN Relation (ours) & \textbf{87.99} & \textbf{79.26} & \textbf{76.33} & \textbf{92.75} & \textbf{89.03} & \textbf{86.24} \\  
         \rowcolor{gray!20}
         Improvement & \textcolor[RGB]{50,205,50}{+0.79} & \textcolor[RGB]{50,205,50}{+0.50} & \textcolor[RGB]{50,205,50}{+0.83} & \textcolor[RGB]{50,205,50}{+0.78} & \textcolor[RGB]{50,205,50}{+1.13} & \textcolor[RGB]{50,205,50}{+1.22} \\
    \bottomrule[1pt]
    \end{tabular}
    }
    \label{tab:res_all_classes_kitti_test}
\end{table}

\subsection{Ablation Studies}
In this section, we analyze the effectiveness of each mechanism in the inter-object relation architecture by extensive ablation experiments. We conduct all model training in ablation studies on the KITTI training set and evaluate them on the validation set. 

\noindent\textbf{Effect of GNN Components.} We compare the effects of different components in the GNN in Tab.~\ref{tab:ablation_study}. We report results of cars on the 3D AP$_{R40}$ metric. We design four-group experiments, including 1) initializing the node features without box information, 2) combining the box information with the proposal features as the initial node features, and only using the feature differences to calculate the edge features and update the node features, 3) concatenating feature differences and proposal box differences to update the node features, and 4) based on the third experiment, concatenating node features of the same node from different layers to be the final output feature of the GNN module. 
\begin{table}[htb]
    \centering
    \caption{Comparison between using different components of relation graph on the KITTI~\cite{geiger2012we} \textbf{validation set}. We report car class (moderate difficulty) using 3D AP$_{R40}$ with IoU threshold 0.7. PV-RCNN~\cite{shi2020pv} is utilized as the basic network. "Init. box" represents using box information to initialize node features and "box diff." is the proposal box differences.}
    \resizebox{\linewidth}{!}{
    \begin{tabular}{ccccc}
    \toprule[1pt]
         Method &  init. box & box diff. & feature appended & 3D AP (Mod.) (\%)~$\uparrow$ \\
         \midrule
         Exp. 1 & - & - & - & 82.91 \\
         Exp. 2 & \checkmark & - & - & 84.07 \\
         Exp. 3 & \checkmark & \checkmark & - & 84.86 \\
         Exp. 4 & \checkmark & \checkmark & \checkmark & \textbf{85.36} \\
    \bottomrule[1pt]
    \end{tabular}
    }
    \label{tab:ablation_study}
\end{table} 
We notice that applying box information to initialize node features improves more than 1\%. The reason is that the original box information provides the spatial state of each proposal, such as position, box size, and rotation angle, which helps the network to effectively capture contextual information and pass messages between the neighbor nodes. We conduct the second experiment with the box differences, which further improves 0.79\% accuracy. This result verifies that learning the similar movement patterns of neighbor objects assists the network in overcoming partial occlusion and predicting more precise bounding box rotation. Additionally, we combine the node features from different GNN layers, making the output features encode various latent features and keeping semantic information. The combination further increases the detection accuracy to 85.36\%.

\noindent\textbf{Effect of Different Graph Generator.}  We explore the influence of different graph generation strategies on the detection performance. We compare two methods, including KNN with k equal to 16 and a radius graph with a range of 6 meters. Usually, fewer proposals are included in the range of 0 to 6 meters in the KITTI dataset. (as shown in Fig.~\ref{fig:graph_example}). In Tab.~\ref{tab:ablation_study_knn_radius}, we illustrate the experiment results of all three categories, such as car, pedestrian, and cyclist, on the moderate difficulty level. The model with the graph generated by KNN outperforms the radius method for cars with 1.43\%. Conversely, the radius model outperforms the KNN model by achieving 1.97\% and 1\% higher accuracy rates for detecting pedestrians and cyclists, respectively. Due to the large distribution range of cars, connecting more neighbor nodes can provide general pattern information to improve performance. In contrast, pedestrians and cyclists tend to cluster densely in scenes; hence, incorporating information from distant objects can introduce noise. However, leveraging proposal information from a relatively short range, which is more pertinent to these proposals, significantly improve the detection of these objects.
\begin{table}[htb]
    \centering
    \caption{Comparison between different graph generation approaches (KNN and Radius) on the KITTI~\cite{geiger2012we} \textbf{validation set}. PV-RCNN~\cite{shi2020pv} is utilzed as the basic network. We set k to 16 and the radius to 6 meters. The evaluation metric is 3D AP$_{R40}$ moderate difficulty with IoU threshold 0.7 for cars and 0.5 for cyclists and pedestrians.}
    \begin{tabular}{ccccc}
    \toprule[1pt]
         \multirow{2}{*}{Method} & \multicolumn{3}{c}{3D AP (Mod.) (\%)~$\uparrow$} \\
         & Car & Pedestrian & Cyclist \\
         \midrule
         PV-RCNN Relation (Radius) & 83.25 & \textbf{59.28} & \textbf{74.29} \\
         PV-RCNN Relation (KNN) & \textbf{84.68} & 57.31 & 73.29 \\
    \bottomrule[1pt]
    \end{tabular}
    \label{tab:ablation_study_knn_radius}
\end{table}

\begin{table}[htb]
    \centering
    \caption{Comparison between the baseline PV-RCNN~\cite{shi2020pv} and our PV-RCNN relation network with various KNN settings on the KITTI~\cite{geiger2012we} \textbf{validation set} for car class using 3D AP$_{R40}$ w.r.t IoU threshold 0.7. Bold and underlined represent each metric's best and second-best performance.}
    \begin{tabular}{ccccc}
    \toprule[1pt]
        \multirow{2}{*}{Method} & \multicolumn{3}{c}{Car - 3D AP (\%)~$\uparrow$} \\
        ~ & Easy & Moderate & Hard \\
        \midrule
        PV-RCNN (baseline) & 91.95 & 84.60 & 82.49 \\
        PV-RCNN Relation (K=16) & \underline{92.04} & \textbf{85.13} & \textbf{82.76}\\
        PV-RCNN Relation (K=32) & \textbf{92.22} & \underline{84.98} & \underline{82.55} \\
    \bottomrule[1pt]
    \end{tabular}
    \label{tab:ablation_study_k_value}
\end{table}

\noindent\textbf{Effect of Various K Values on KNN-based Graph.} Moreover, we explore the impact of k values on the KNN-based graph generator for the performance of 3D object detection. We compare the baseline model with our object relation approach using different k values (16 and 32). We show the experiment results based on the PV-RCNN~\cite{shi2020pv} network in Tab.~\ref{tab:ablation_study_k_value}. When k is 16, the PV-RCNN relation model achieves the best performances for the moderate and hard-level cars. If we utilize a larger value (k=32), the object relation module performs the best on easy difficulty (92.22\%), while the 3D AP of moderate and hard levels decreases. The reason is that a suitable k value helps the network extend the receptive field, assisting the detector in perceiving the surrounding environment. However, a larger k value leads the relation module to establish a connection with more objects, which can introduce noise into the object features, causing a decrease in the object detection performance.

\subsection{Discussion and Limitations}

\noindent\textbf{Stability and Reliability.}
\begin{table}[]
    \centering
    \caption{Stability Comparison between the baseline PV-RCNN~\cite{shi2020pv} and ours on the KITTI~\cite{geiger2012we} \textbf{validation set}. Trained and evaluated only on the car class. Both models were trained three times and the average is reported. We report 3D AP$_{R11}$ and 3D AP$_{R40}$ with IoU threshold 0.7. Frame rate indicates frames per second w.r.t inference speed.}
    \resizebox{\linewidth}{!}{
    \begin{tabular}{ccccc}
        \toprule[1pt]
         \multirow{2}{*}{Method} & Frame & Easy & Moderate & Hard \\
         & rate & R11/R40 & R11/R40 & R11/R40 \\
         \midrule
         PV-RCNN & \textbf{5.8} & 89.39/92.02 & 83.63/84.80 & 78.86/82.58 \\
         PV-RCNN Relation (ours) & 5.6 & \textbf{89.59/92.53} & \textbf{84.56/85.22} & \textbf{79.04/82.99} \\
         \rowcolor{gray!20}
         Improvement & \textcolor{red}{-0.2} & \textcolor[RGB]{50,205,50}{+0.20}/\textcolor[RGB]{50,205,50}{+0.51} & \textcolor[RGB]{50,205,50}{+0.93}/\textcolor[RGB]{50,205,50}{+0.42} & \textcolor[RGB]{50,205,50}{+0.18}/\textcolor[RGB]{50,205,50}{+0.41} \\
         \bottomrule[1pt]
    \end{tabular}
    }
    \label{tab:stability_reliability}
\end{table}
We repeat all experiments with same settings three times to confirm the stability and reliability of our method. The average results are presented in Tab.~\ref{tab:stability_reliability}. Our relational approach consistently enhances 3D object detection performance under both AP$_{R11}$ and AP$_{R40}$ metrics, while adding only a tiny inference time. Notably, accuracy improvements exceeding 0.4\% for moderate and hard difficulty levels demonstrate the efficacy of our inter-object relational approach in complex driving conditions.

\begin{table}[htb]
    \centering
    \caption{Performance Comparison between GACE~\cite{schinagl2023gace} and ours on the KITTI~\cite{geiger2012we} \textbf{validation set}. We present the 3D AP$_{R40}$ results for car, pedestrian, and cyclist under moderate difficulty. While PV-RCNN$^{*}$ is reported by~\cite{schinagl2023gace}, PV-RCNN is a self-trained version.}
    \resizebox{\linewidth}{!}{
    \begin{tabular}{ccccc}
        \toprule[1pt]
        \multirow{2}{*}{Method} & \multirow{2}{*}{Reference} & \multicolumn{3}{c}{3D AP (Mod.) (\%) $\uparrow$}\\
        & & Car & Pedestrian & Cyclist \\
        \midrule
        PV-RCNN$^{*}$~\cite{schinagl2023gace} &  & 82.86 & 53.64 & 70.42 \\
        PV-RCNN GACE~\cite{schinagl2023gace} & ICCV23 & 82.84 & 61.06 & 72.70 \\
        \rowcolor{gray!20}
        Improvement & & \textcolor{red}{-0.02} & \textcolor[RGB]{50,205,50}{+7.42} & \textcolor[RGB]{50,205,50}{+2.28} \\
        \midrule
        PV-RCNN & & 84.53 & 57.99 & 70.81 \\
        PV-RCNN Relation (ours) & & 85.36 & 58.13 & 71.13 \\
        \rowcolor{gray!20}
        Improvement & & \textcolor[RGB]{50,205,50}{+0.83} & \textcolor[RGB]{50,205,50}{+0.14} & \textcolor[RGB]{50,205,50}{+0.32} \\
        \bottomrule[1pt]
    \end{tabular}
    }
    \label{tab:improvement_comparison}
\end{table}
\noindent\textbf{Comparison with Other Graph-based Approach.}
In Tab.~\ref{tab:improvement_comparison}, we compare our approach with a state-of-the-art approach, GACE~\cite{schinagl2023gace}, which improves detection confidence by using final prediction results in the near range. GACE also utilizes points within each predicted bounding box to provide local features, achieving significant improvements for pedestrians and cyclists compared to its baseline. However, it fails to improve detection performance of cars. Conversely, our approach shows fewer improvements for pedestrians and cyclists but demonstrates consistent effectiveness across all classes.

Our inter-object relation module effectively learns and utilizes the relationship and movement patterns among various objects. Nevertheless, our method does not fully exploit point-level features, which is a potential reason for the limited improvements for pedestrians and cyclists. Because cyclists and pedestrians usually appear small and contain only a few points. Relying solely on global object relations from proposal levels alleviates occlusion but does not yield precise information about the status of these small objects. Hence, integrating point-level features is crucial for further improving the performance of our relation approach.


\section{Conclusion and Future Works}
In this study, we present an inter-object relation module to improve the performance of arbitrary two-stage detectors in 3D object detection. This module leverages a graph neural network to extract features from inter-object relationship graphs, refining proposal features to improve detection accuracy in occlusion and distant cases. We exploit specific patterns, such as parallel parking on narrow streets, to refine the predicted object direction.viso Our evaluation on the KITTI dataset demonstrates improvements of 0.58\% and 0.47\% in detecting cars at the hard difficulty level over the PV-RCNN and PartA$^2$ baseline models, respectively. Furthermore, our method significantly outperforms the baseline on the KITTI official leaderboard. We further conduct comprehensive ablation studies to verify the effectiveness of core components of our proposed inter-object relation approach. Future works include evaluating our approach on other large-scale datasets and extending it to more two-stage networks.
\label{conclusion}

{
    \bibliographystyle{ieeetr}
    \bibliography{reference}
}

\end{document}